\documentclass{article}

\usepackage{arxiv}

\usepackage{amsmath,amsfonts}
\usepackage{array}
\usepackage{textcomp}
\usepackage{stfloats}
\usepackage{url}
\usepackage{verbatim}
\usepackage{graphicx}
\usepackage{cite}

\usepackage[colorlinks,urlcolor=blue,linkcolor=blue,citecolor=blue]{hyperref}
\usepackage{color,array}
\usepackage{balance}
\usepackage{multirow}%
\usepackage{amsmath,amssymb,amsfonts}%
\usepackage{amsthm}%
\usepackage{mathrsfs}%
\usepackage{xcolor}%
\usepackage{textcomp}%
\usepackage{manyfoot}%
\usepackage{booktabs}%
\usepackage{algorithm}%
\usepackage{algorithmicx}%
\usepackage{algpseudocode}%

\usepackage{listings}%
\usepackage{adjustbox}
\usepackage{float}
\usepackage{csquotes}
\usepackage{xurl}
\usepackage{graphicx}
\usepackage{multicol}
\usepackage{comment}
\usepackage{caption}
\usepackage{subcaption}
\usepackage[table]{xcolor}  
\usepackage{threeparttable} 
\usepackage{svg} 

\definecolor{deepblue}{RGB}{243, 252, 255}
\definecolor{lightblue}{RGB}{230, 243, 247}

\title{Neural Residual Modeling for Scientific Data Compression under Guaranteed Error Bounds}

\author{ 
    Surya Majumder, Liangji Zhu, Sanjay Ranka and Anand Rangarajan \\ \\
	Department of Computer \& Information Science \& Engineering \\
	University of Florida, Gainesville, FL, USA \\
}

\date{}

\renewcommand{\headeright}{}
\renewcommand{\undertitle}{}
\renewcommand{\shorttitle}{Neural Residual Modeling for Scientific Data Compression}

\hypersetup{
pdftitle={Neural Residual Modeling for Scientific Data Compression under Guaranteed Error Bounds},
pdfauthor={Surya Majumder, Liangji Zhu, Sanjay Ranka, Anand Rangarajan},
pdfkeywords={scientific data compression, neural networks, residual modeling, error bounds, lossy compression},
}

\begin{document}
\maketitle

\begin{abstract}
	Lossy compression of scientific simulation data increasingly relies on learned, latent-space architectures such as Residual Vector Quantization (RVQ), which iteratively quantize a base representation and its residuals to progressively reduce reconstruction error. While effective, RVQ performs this residual modeling entirely in latent space, leaving the pixel-space error structure of the reconstruction largely unaddressed. In this work, we propose a post-processing pipeline that augments an RVQ-based compressor with a U-Net trained to predict and correct pixel-space residuals between the original volume and its RVQ reconstruction. We show that these residuals are spatially structured rather than driven by local intensity or gradient features, motivating the need for a deep spatial model rather than simple statistical correction. The U-Net-corrected reconstruction is then passed through a Guaranteed Autoencoder (GAE) stage, which projects the remaining residual onto a per-block PCA basis to enforce a user-specified block-wise error bound. To the best of our knowledge, this is the first pipeline to combine latent-space RVQ, explicit pixel-space residual correction via a deep spatial post-processing network, and GAE-based error-bound guarantees within a single framework for scientific data compression. We evaluate our approach on S3D, JHTDB and E3SM datasets, demonstrating consistent improvements in NRMSE, compression ratio] over RVQ-only and standard residual-correction baselines, while maintaining strict error guarantees required for scientific data fidelity.
\end{abstract}


\section{Introduction}
The explosive growth of high-resolution scientific simulations, ranging from climate modeling to fluid dynamics, cosmology, and fusion energy research, has resulted in data volumes that far outpace the growth of storage, network bandwidth, and I/O capabilities \cite{ainsworth2018multilevel}. A single simulation run can produce tens of terabytes of 3D volumetric data across thousands of timesteps, making it infeasible to store or transfer such data in its raw form. Consequently, lossy compression has emerged as an essential tool for scientific data management, enabling substantial reductions in data size while preserving the fidelity required for downstream scientific analysis \cite{gong2023mgard}.

Traditional error-bounded lossy compressors, such as SZ and ZFP, rely on hand-crafted predictors and transform-based encoding to guarantee point-wise error bounds and block-wise error bounds respectively \cite{liang2022sz3}, \cite{lindstrom2014fixed}. While effective and widely adopted, these methods are inherently limited by the expressiveness of their fixed predictors, which struggle to capture the complex, nonlinear spatial correlations present in scientific data. This has motivated a growing body of work on learned compressors, which use neural networks, particularly autoencoders, to learn data-driven representations that can adapt to the underlying structure of the data \cite{balle2018variational}, \cite{li2025caesar}.

Among these learned approaches, Residual Vector Quantization (RVQ) has gained traction as an effective mechanism for progressive, high-fidelity compression. RVQ operates by quantizing a base latent representation and then iteratively quantizing the residual error between the original latent and its quantized approximation, with each additional quantization stage further reducing the reconstruction error \cite{razavi2019generating}. This residual refinement strategy allows RVQ-based compressors to achieve fine-grained control over the rate-distortion trade-off by simply adjusting the number of quantization stages.
However, a key limitation of RVQ, and indeed of latent-space compression more broadly, is that all residual modeling occurs within the latent space of the encoder. The residual that RVQ corrects is the difference between a latent vector and its quantized codeword, not the difference between the original data and its final pixel-space reconstruction. As a result, even after all RVQ stages are applied, the decoded reconstruction can retain a pixel-space error that, while implicitly minimized during training via reconstruction loss, is never explicitly modeled or corrected as a structured spatial signal after decoding. This raises a natural question: if the goal is ultimately to minimize reconstruction error in the original data domain, why should residual correction be confined to the latent space at all?

To investigate this gap, we analyze the pixel-space residuals left behind after RVQ-based reconstruction. Through correlation analysis between these residuals and local features such as pixel intensity and gradient magnitude, we find that the residual error is not primarily driven by simple local statistics. Instead, the error exhibits spatial structure that is not easily summarized by per-pixel intensity or edge information alone. This finding suggests that no simple, intensity-based correction scheme can adequately capture the residual error pattern; instead, a model capable of learning spatial dependencies is required. 

Motivated by this observation, we propose a post-processing residual correction stage built on a U-Net architecture, applied directly to the pixel-space output of an RVQ-based compressor. Rather than modifying the RVQ pipeline itself, our U-Net operates as a lightweight, modular add-on: it takes the RVQ reconstruction as input, predicts the pixel-space residual between this reconstruction and the original data, and adds this predicted residual back to produce a corrected reconstruction. Because the U-Net is trained specifically to model the spatial error structure left behind by RVQ, it is able to recover a substantial portion of the reconstruction error that RVQ's latent-space residual modeling does not address, without requiring any retraining or architectural changes to the underlying compressor.

While this U-Net correction step significantly improves reconstruction quality, scientific applications often demand strict, guaranteed point-wise error bounds, a requirement that learned models alone cannot satisfy, since their training objectives minimize aggregate error rather than bound any individual value. To close this gap, we apply a Guaranteed Autoencoder (GAE) stage \cite{li2025machine, lee2023nonlinear} as the final step of our pipeline. GAE operates by computing the remaining residual between the original data and the U-Net-corrected reconstruction, projecting this residual onto a per-block SVD basis, and retaining only as many coefficients as are needed to bring the per-patch $\ell_2$ norm of the residual below a user-specified NRMSE tolerance $\tau$. This final stage guarantees that, regardless of how well the RVQ and U-Net stages perform, the overall pipeline satisfies the required error bound.

Taken together, our contributions are as follows:
\begin{itemize}

\item We identify and analyze a fundamental limitation of RVQ-based compression: residual modeling is performed entirely in latent space, leaving pixel-space reconstruction errors unaddressed.

\item We show, through correlation analysis, that these pixel-space residuals are spatially structured rather than driven by local intensity or gradient features, motivating the need for a deep spatial model for residual correction.

\item We propose a modular, post-processing U-Net that operates directly on RVQ reconstructions to predict and correct pixel-space residuals, improving reconstruction fidelity without modifying the underlying compressor.

\item We integrate a Guaranteed Autoencoder (GAE) stage to enforce strict, user-specified block-wise error bounds on the final corrected reconstruction.

\item To the best of our knowledge, this is the first pipeline to combine latent-space RVQ, pixel-space residual correction via a deep post-processing network, and GAE-based error guarantees within a single framework for scientific data compression.

\end{itemize}
The remainder of this paper is organized as follows. \autoref{sec:related} reviews related work in error-bounded lossy compression and learned compression methods. \autoref{sec:method} describes our proposed pipeline in detail. \autoref{sec:results} presents experimental results across the datasets. \autoref{sec:conclusion} concludes the paper and discusses future directions.

\section{Related Work}\label{sec:related}

\subsection{Algorithmic Error-Bounded Compression}
Error-bounded lossy compression of scientific floating-point data has traditionally relied on algorithmic, prediction-based techniques that require no training data. The SZ family of compressors \cite{liang2022sz3} predicts each grid point from its previously decoded neighbors using a combination of linear regression, interpolation, and the Lorenzo predictor \cite{ibarria2003out}, then quantizes and entropy-codes the prediction residual under a user-specified error bound. ZFP \cite{lindstrom2014fixed} instead partitions data into fixed-size blocks and applies a near-orthogonal transform for decorrelation, primarily targeting fixed-rate or fixed-precision compression rather than strict per-point error bounds; while SPERR \cite{li2023lossy} uses wavelet transforms with SPECK coefficient coding. These methods share a common structure: a fixed, non-adaptive predictor estimates each value, and only the residual between the prediction and the true value is stored, with the residual's magnitude directly controlling the compression ratio at a given error tolerance.

While fast and interpretable, algorithmic compressors are fundamentally limited by the expressiveness of their predictors. The Lorenzo predictor, for instance, is purely linear and performs well on smooth fields but degrades sharply near shocks, sharp gradients, or other discontinuities, where its prediction error grows and more bits are required to encode the residual. This limitation has motivated a growing body of work on learned, neural compressors that can adapt to the nonlinear structure of scientific data.

\subsection{Learned Scientific Data Compression}
Variational autoencoders (VAEs) with hyperprior structures \cite{balle2018variational} have become the dominant architecture for learned compression, encoding data into a compact latent representation whose statistics are modeled by a learned entropy model. AE-SZ \cite{liu2021exploring} was among the first to bring this idea into the scientific domain, using a convolutional autoencoder as an additional predictor inside the SZ pipeline: for each block, the autoencoder's prediction is compared against the Lorenzo predictor's, and the better of the two is selected, with the residual quantized under the target error bound. SRN-SZ \cite{liu2023srn} takes a different approach, training a super-resolution network to reconstruct full-resolution data from a sparsified coarse grid, with the SR network learning the full mapping from coarse to fine rather than a residual correction.

The base compressor, CAESAR-V \cite{li2025caesar}, extends the hyperprior VAE framework to 3D scientific data by alternating 2D and 3D convolutions to capture spatiotemporal correlations, combined with a super-resolution module at the decoder. Crucially, CAESAR-V performs a single forward pass through its encoder-decoder pipeline to produce a pixel-space reconstruction $\tilde{x}$; it does not include any mechanism for explicitly modeling or correcting the residual $r = x - \tilde{x}$ between this reconstruction and the original data. This is the gap our pipeline addresses.

\subsection{Latent-Space Residual Quantization}
Residual Vector Quantization (RVQ) \cite{razavi2019generating} is a cascaded quantization scheme originally developed for neural audio codecs \cite{defossez2022high, zeghidour2021soundstream}, in which a sequence of $N$ codebooks iteratively quantizes the residual left by the previous stage: given a continuous latent $z$, the first codebook produces $\hat{z}_1 = \text{Quantize}(z)$, the second quantizes $z - \hat{z}_1$, and so on, with the final quantized latent given by $\hat{z} = \sum_{k=1}^{N} \hat{z}_k$. Each additional stage reduces the quantization error in the latent representation, allowing fine-grained control over the rate-distortion trade-off.

RVQ operates entirely \emph{before} decoding, on the continuous latent representation $z$ produced by the encoder. It reduces the discretization error introduced when $z$ is mapped to a finite codebook, but it has no access to the pixel-space reconstruction $\tilde{x} = \text{Decoder}(\hat{z})$, nor to the spatial structure of the error $x - \tilde{x}$ that remains after decoding. In other words, RVQ is a residual modeler only in the narrow sense of correcting quantization error in latent space; it does not perform spatial residual learning in the data domain, and the residual it corrects is not the same residual that determines the final reconstruction quality of the decoded output.

This distinction motivates the central question of our work: if the ultimate objective is to minimize the pixel-space reconstruction error $\|x - \tilde{x}\|$, why should residual correction be confined to the latent space at all?

\subsection{Pixel-Space Residual Learning and Decoder-Side Post-Processing}
A separate line of work, originating in image restoration, trains neural networks to predict the residual between a degraded observation and a clean target, then adds this prediction back to obtain the corrected output. DnCNN \cite{zhang2017beyond} established this paradigm for image denoising, showing that training a network to predict the noise $v = x - x'$ (rather than the clean image $x$ directly) is both more stable and more accurate, since the residual has simpler statistics than the image itself. The same residual-target principle underlies decoder-side enhancement modules in learned image compression \cite{liu2020unified}, artifact-removal networks for JPEG and BPG-compressed images \cite{kirmemics2018learned}, and recurrent residual coding schemes \cite{toderici2017full, johnston2018improved}.

Most closely related to the architecture we propose is NU-Class Net \cite{zilouchian2024nu}, which uses a U-Net \cite{ronneberger2015u} to predict the residual between a video codec's output and the original frame, then adds the predicted residual back to the decoded frame. This demonstrates that a full encoder-decoder network with skip connections, applied after decoding, can effectively model the spatial structure of compression artifacts. However, this work operates on natural video for perceptual quality, with no error-bound guarantee and no application to 3D scientific volumetric data.

Deep Lossy Plus Residual (DLPR) coding \cite{bai2024deep} is structurally the closest analog in the natural-image literature to our overall pipeline shape: a VAE produces a lossy reconstruction $\tilde{x}$, and the residual $r = x - \tilde{x}$ is then handled by a second stage. The key difference is that DLPR's second stage \emph{compresses} the residual using a learned entropy coder, storing it explicitly; it does not \emph{predict} the residual from $\tilde{x}$ alone. Our U-Net instead learns a mapping $\tilde{x} \mapsto \hat{r} \approx r$ without access to $r$ itself at inference time, a harder prediction task that avoids the additional bitrate cost of an explicit residual stream.

\subsection{Neural Residual Enhancement for Scientific Compressors}
The work closest in spirit and domain to ours is NeurLZ \cite{jia2025neurlz}, which attaches a lightweight neural network (on the order of 3{,}000 parameters) to a traditional compressor (SZ3 or ZFP), training it online at compression time to predict the residual between the compressor's output and the original data. The predicted residual is stored and added back at decompression. A related extension, GWLZ \cite{jia2024gwlz}, partitions the data into groups with distinct statistical regimes and trains a separate lightweight model per group, reflecting the observation that scientific fields contain spatially heterogeneous regions whose residuals follow different distributions.

While NeurLZ and GWLZ share our core strategy of predicting a pixel-space residual and adding it back to a compressor's output, three differences are central to our contribution. First, their base compressor is a traditional predictor-quantization compressor (SZ3 or ZFP), whereas ours is a learned VAE; the residual left behind by a VAE has substantially different structure than that left by a linear predictor, since the VAE has already performed nonlinear dimensionality reduction. Second, their residual predictors are small, flat networks operating on 2D slices, whereas our U-Net is a full encoder-decoder with skip connections, capable of modeling multi-scale spatial structure per frame across the volume. Third, neither method includes a dedicated downstream per-patch NRMSE guarantor separate from the base compressor; NeurLZ and GWLZ inherit the error control of SZ3 at the compressor level, but the neural residual correction layer itself carries no independent error certificate; our pipeline couples the U-Net correction with a Guaranteed Autoencoder (GAE) stage \cite{li2024attention, lee2022error} that certifies the final output meets a user-specified NRMSE bound regardless of how well the upstream stages perform.

\subsection{Guaranteed Error Bounds via PCA-Based Correction}
The Guaranteed Autoencoder (GAE) framework \cite{li2024attention} enforces per-block error guarantees by computing the residual between a learned reconstruction and the original data, projecting this residual onto a per-block PCA basis, and retaining the minimum number of coefficients needed to bring the residual's $\ell_2$-norm below a target threshold $\tau$. Because the retained coefficients are computed exactly from the residual, this approach provides a hard guarantee, in contrast to the aggregate, average-case error reduction that a neural network's training objective alone can offer. Related formulations have used the same PCA-projection principle to preserve derived quantities of interest in climate data \cite{lee2022error} and to certify error bounds for compressive autoencoders more generally \cite{lee2023nonlinear}.

A closely related recent direction, Guaranteed Conditional Diffusion (GCDTC) \cite{lee2025guaranteed}, also combines a U-Net-based architecture with a GAE-style error-bound enforcer for 3D scientific data. The critical distinction is architectural placement: in GCDTC, the U-Net is the conditional diffusion \emph{decoder} itself, operating in latent space during the iterative denoising process that produces the reconstruction. In our pipeline, the U-Net is a separate, dedicated post-decoder module that operates entirely in pixel space on the already-decoded reconstruction $\tilde{x}$, with the explicit goal of predicting and correcting the residual $r = x - \tilde{x}$ before the GAE stage is invoked.

Most recently, Zhu et al. \cite{zhu2026residual} proposed two complementary residual coders that specifically target the high-fidelity regime where GAE-style global PCA correction becomes rate-dominant. Their first method, LBRC (Lorenzo-Based Residual Coding), is a training-free pipeline that adaptively quantizes the learned residual to a target NRMSE, then losslessly encodes the resulting integer residual using 3D Lorenzo differencing, zigzag mapping, bit-plane coding, and entropy coding. Their second method, NGLR (Neural-Guided Lorenzo Residual Coding), extends LBRC with a lightweight causal neural bias predictor that uses features from the base reconstruction and already-decoded quantized residuals to correct systematic errors in the Lorenzo prediction, reducing the entropy of the residual code stream while preserving fully deterministic decoding. Across E3SM, JHTDB, and ERA5 at block-level NRMSE targets between $10^{-6}$ and $10^{-4}$, LBRC improves compression ratio over GAE by 30--60\% and NGLR adds a further 10--40\% over LBRC. The key distinction from our work is both directional and regime-specific: LBRC and NGLR replace the GAE correction stream with a more efficient residual representation, whereas our U-Net stage reduces the magnitude of the residual that the GAE must correct in the first place, making fewer PCA coefficients sufficient to meet the bound. The two approaches are therefore complementary. Our pipeline operates in the moderate-fidelity regime where the base reconstruction dominates and the GAE correction cost is manageable, while LBRC and NGLR target the high-fidelity regime where the correction stream itself becomes the bottleneck.

Each individual component of our pipeline has precedent: VAE-based learned compression with RVQ-style latent refinement is well established \cite{li2025caesar, razavi2019generating}; residual-target supervision for pixel-space correction is foundational in image restoration \cite{zhang2017beyond, he2016deep}; U-Net architectures for residual prediction after decoding have been demonstrated in video \cite{zilouchian2024nu}; and PCA-based guaranteed error bounds are established for scientific data \cite{li2024attention}. However, no prior work combines a learned VAE base compressor, a dedicated full U-Net operating in pixel space to predict the reconstruction residual, followed by a downstream GAE error-bound enforcer into a single pipeline for scientific volumetric data. NeurLZ \cite{jia2025neurlz} and GWLZ \cite{jia2024gwlz} are the closest published methods, but both operate on traditional compressors with small, flat networks and lack an error-bound guarantee stage; NU-Class Net \cite{zilouchian2024nu} demonstrates the U-Net residual-prediction paradigm but in the natural-video domain without error bounds; and GCDTC \cite{lee2025guaranteed} pairs a U-Net with GAE but places the U-Net as the decoder itself rather than as a post-decoder corrector. Our work is, to the best of our knowledge, the first to occupy this specific position in the pipeline for scientific data compression.

\section{Proposed Methodology}
\label{sec:method}

\begin{figure}[!htbp]
    \centering
    \includegraphics[width=\linewidth]{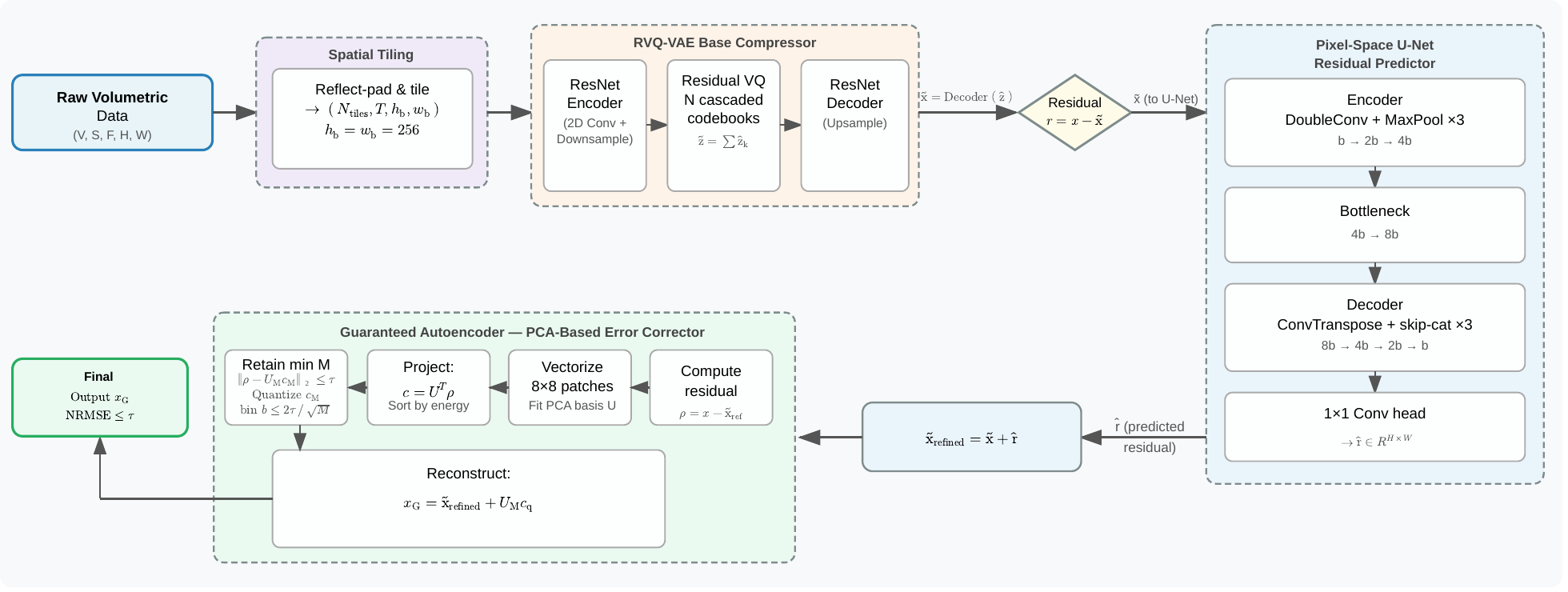}
    \caption{Overview of the proposed three-stage compression pipeline. The input volume is tiled into spatial blocks and passed through an RVQ-based VAE (Stage 1) to obtain a pixel-space reconstruction $\tilde{x}$. A pixel-space U-Net (Stage 2) predicts the residual $\hat{r} = x - \tilde{x}$ and produces a refined reconstruction $\tilde{x}_{\text{refined}} = \tilde{x} + \hat{r}$. A Guaranteed Autoencoder (Stage 3) then projects the remaining residual onto a per-block PCA basis to enforce a user-specified NRMSE bound $\tau$, producing the final certified output $x^G$.}
    \label{fig:pipeline}
\end{figure}

Our pipeline augments an RVQ-based base compressor with two modular post-processing stages: a pixel-space U-Net that explicitly predicts and corrects the reconstruction residual, followed by a Guaranteed Autoencoder (GAE) stage that enforces a hard per-block NRMSE bound on the final output. \autoref{fig:pipeline} provides an overview of the complete pipeline.

\subsection{Problem Formulation}

Let $x \in \mathbb{R}^{H \times W}$ denote a single spatial frame of a volumetric scientific dataset, and let $\tau > 0$ denote a user-specified error tolerance. The goal of error-bounded lossy compression is to produce a reconstruction $x^G$ such that
\begin{equation}
    \mathrm{NRMSE}(x, x^G) = \frac{\sqrt{\frac{1}{n}\|x - x^G\|_2^2}}{x_{\max} - x_{\min}} \leq \tau
    \label{eq:nrmse}
\end{equation}
while minimizing the total bitrate required to represent $x^G$. Our pipeline decomposes this objective into three complementary stages, each targeting a distinct source of reconstruction error.

\subsection{Data Preprocessing and Spatial Tiling}

Volumetric scientific datasets arrive with shape $(V, S, F, H, W)$, where $V$ denotes the number of variables, $S$ the number of sections, $F$ the number of frames, and $H \times W$ the spatial resolution of each frame. Since operating on the full spatial extent of a frame is computationally prohibitive for both the U-Net and the GAE, we first tile each frame into non-overlapping spatial blocks of size $h_b \times w_b$ (set to $256 \times 256$ in our experiments). If $H$ or $W$ is not divisible by the block size, we apply reflect padding before tiling and crop it off after reassembly, ensuring that no information is discarded at boundaries.

Each frame is jointly normalized using the global mean $\mu$ and range $\Delta = x_{\max} - x_{\min}$ of the original data, mapping values to $[0, 1]$:
\begin{equation}
    x^{(01)} = \frac{(x - \mu)/\Delta - \ell_{\text{joint}}}{r_{\text{joint}}}
    \label{eq:norm}
\end{equation}
where $\ell_{\text{joint}}$ and $r_{\text{joint}}$ are the joint minimum and range computed across both the original and reconstructed data, ensuring the U-Net input and target occupy the same numerical range. This normalization is fully invertible and is reversed before the final output is written.

\subsection{Stage 1: RVQ-Based Latent Compression}

The first stage of our pipeline is a convolutional autoencoder with Residual Vector Quantization (RVQ) as the discrete bottleneck, which we refer to as the RVQ autoencoder; building directly on the CAESAR-V architecture \cite{li2025caesar}. The encoder $E_\phi$ maps each spatial tile $x \in \mathbb{R}^{h_b \times w_b}$ to a continuous latent representation $z = E_\phi(x)$ through a sequence of ResNet blocks and stride-2 downsampling layers, progressively reducing spatial resolution while increasing channel depth.

The continuous latent $z$ is then quantized using $N$ cascaded codebooks $\{\mathcal{B}^k\}_{k=1}^N$, each of size $K$. The RVQ quantization proceeds iteratively: the first codebook quantizes $z$ to $\hat{z}_1 = \arg\min_{e \in \mathcal{B}^1} \|z - e\|_2^2$; each subsequent stage $k$ quantizes the residual left by all previous stages:
\begin{equation}
    \hat{z}_k = \arg\min_{e \in \mathcal{B}^k} \bigl \| z - \sum_{i=1}^{k-1} \hat{z}_i - e \bigr \|_2^2.
    \label{eq:rvq}
\end{equation}
The final quantized latent is $\hat{z} = \sum_{k=1}^{N} \hat{z}_k$, and the decoder $D_\psi$ produces the pixel-space reconstruction $\tilde{x} = D_\psi(\hat{z})$.

At compression time, the encoder produces the sequence of codebook indices $\{k_1^{(j)}, \ldots, k_N^{(j)}\}$ for each spatial position $j$, where $k_n^{(j)} \in \{1, \ldots, K\}$ identifies the nearest codebook entry at stage $n$. These indices, along with the codebook vectors themselves, constitute the compressed representation. The indices are entropy-coded using Huffman coding based on their empirical frequency distribution. At decompression time, the decoder reconstructs $\hat{z} = \sum_{k=1}^{N} \hat{z}_k$ from the stored indices and codebook vectors, then passes $\hat{z}$ through the decoder $D_\psi$ to produce $\tilde{x}$. The total bitstream for Stage 1 therefore consists of the Huffman-coded index stream and the stored codebook vectors across all $N$ stages, as detailed in \autoref{tab:bitstream}.

Each stage of RVQ reduces the \emph{quantization error in latent space}, i.e., the discrepancy between the continuous latent $z$ and its discrete approximation $\hat{z}$. However, RVQ operates entirely before decoding: it never observes the pixel-space reconstruction $\tilde{x}$, nor the pixel-space residual $r = x - \tilde{x}$ that persists after the decoder is applied. So, while the RVQ is trained to minimize this pixel-space error in expectation, the residual $r = x - \tilde{x}$ that remains at inference time is never explicitly modeled or corrected as a structured spatial signal after decoding. It is this post-decoding structured residual that our U-Net stage is designed to address.

The RVQ is trained end-to-end with a combined mean squared error and commitment loss
\begin{equation}
    \mathcal{L}_{\text{RVQ}} = \text{MSE}(x, \tilde{x}) + \lambda \cdot \mathcal{L}_{\text{commit}}
    \label{eq:rvqloss}
\end{equation}
where $\mathcal{L}_{\text{commit}} = \sum_{k=1}^{N} \|\text{sg}[z - \sum_{i<k}\hat{z}_i] - \hat{z}_k\|_2^2$ is the commitment loss summed across all quantization stages, $\text{sg}[\cdot]$ denotes the stop-gradient operator, and $\lambda$ controls the trade-off between reconstruction fidelity and codebook commitment stability. Codebook updates use exponential moving average (EMA) with decay $\gamma = 0.99$ and dead-code restart to prevent codebook collapse.

Unlike variational autoencoders that learn a continuous latent distribution with a rate term, our model uses a purely discrete bottleneck: the encoder maps each tile to a continuous latent $z$, which is then quantized to a sum of codebook vectors via the RVQ cascade. The compression ratio is determined by the entropy of the resulting codebook indices, which are Huffman-coded at compression time and codebook size for the RVQ.

\subsection{Stage 2: Pixel-Space U-Net Residual Predictor}

\begin{figure}[!htbp]
    \centering
    \includegraphics[width=\linewidth]{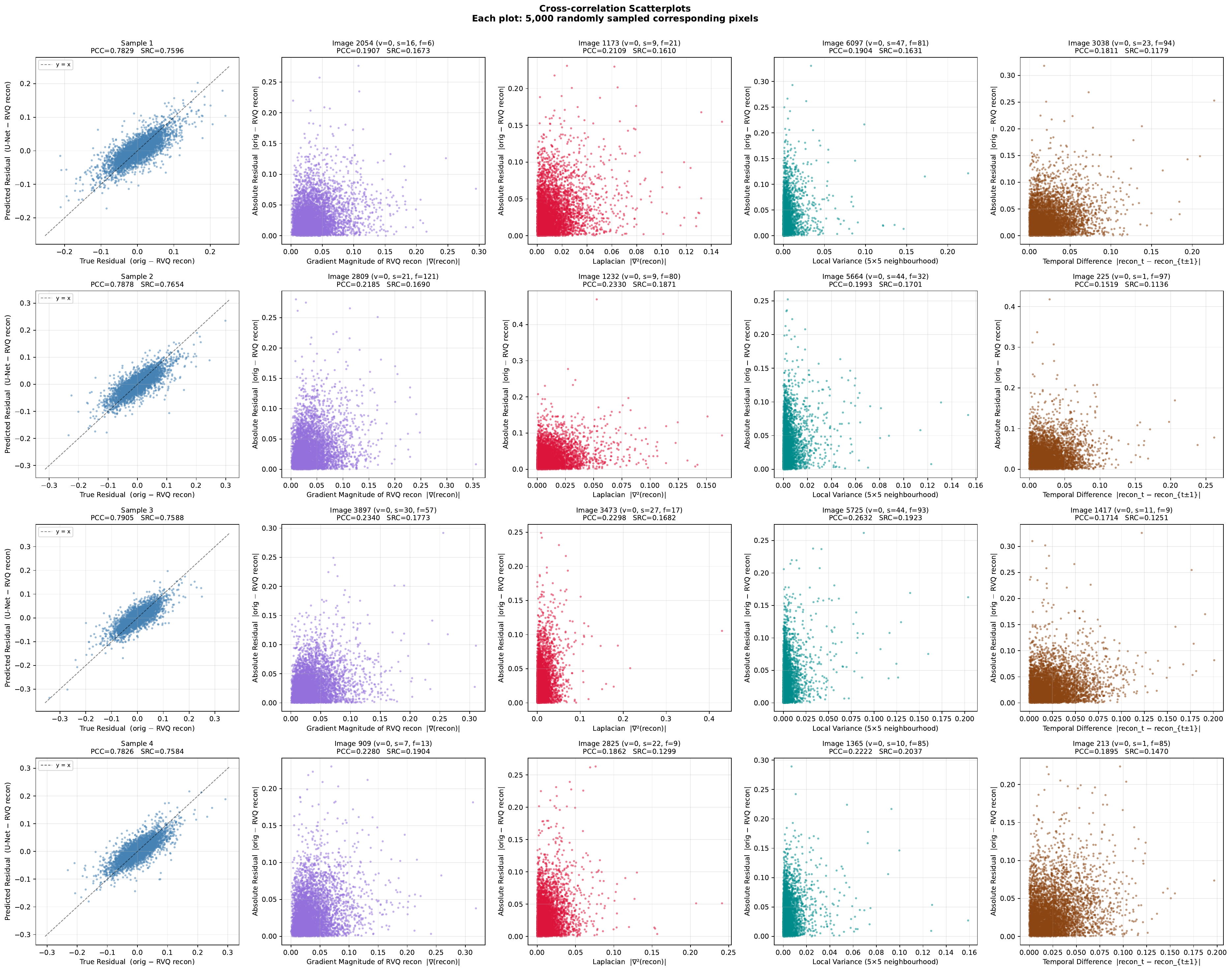}
    \caption{Correlation between RVQ pixel-space residual magnitude and local structural features on the JHTDB dataset. Scatter plots show residual magnitude against gradient magnitude (column 2), Laplacian response (column 3), local variance (column 4), and temporal difference (column 5), each computed per-pixel. The low correlation across all four features indicates that no simple pixel-wise or gradient-wise statistic explains where RVQ reconstruction fails, motivating a deep spatial model capable of learning the more complex underlying pattern. This is further highlighted by the high correlation between the true and predicted residuals from the U-Net (column 1) showing that the U-Net is able to capture some spatial features resulting in proper residual predictions.}
    \label{fig:residual_corr}
\end{figure}

After Stage 1, the RVQ-decoded reconstruction $\tilde{x}$ retains a residual $r = x - \tilde{x}$ that has never been directly modeled. We analyze this residual on the datasets, computing its correlation with local structural features including gradient magnitude, Laplacian response, local variance, and temporal difference, as shown in \autoref{fig:residual_corr} for the JHTDB dataset. We find that the residual is \emph{not} primarily driven by simple intensity-based statistics. Instead, it exhibits spatially structured, multi-scale patterns that cannot be captured by any per-pixel or per-patch statistical correction. This motivates the use of a full encoder-decoder network with skip connections, specifically a U-Net, as the residual predictor.

\subsubsection{Architecture}

Our U-Net takes the RVQ reconstruction $\tilde{x} \in \mathbb{R}^{1 \times h_b \times w_b}$ as input and outputs a predicted residual $\hat{r} \in \mathbb{R}^{1 \times h_b \times w_b}$. The architecture consists of three encoder levels, a bottleneck, and three symmetric decoder levels, with skip connections concatenating encoder feature maps to the corresponding decoder level.

Let $b$ denote the base channel count. Each level of the encoder applies a \texttt{DoubleConv} block — two sequential $3 \times 3$ convolutions with batch normalization and GELU activation, followed by $2 \times 2$ max-pooling:
\begin{align}
    e_1 &= \text{DoubleConv}(\tilde{x},\ b) \notag \\
    e_2 &= \text{DoubleConv}(\text{Pool}(e_1),\ 2b) \notag \\
    e_3 &= \text{DoubleConv}(\text{Pool}(e_2),\ 4b) \notag \\
    b_\text{neck} &= \text{DoubleConv}(\text{Pool}(e_3),\ 8b).
    \label{eq:unet_enc}
\end{align}

The decoder progressively restores spatial resolution using transposed convolutions, concatenating the corresponding encoder map at each level (skip connections):
\begin{align}
    d_3 &= \text{DoubleConv}([\text{Up}(b_\text{neck}),\ e_3],\ 4b) \notag \\
    d_2 &= \text{DoubleConv}([\text{Up}(d_3),\ e_2],\ 2b) \notag \\
    d_1 &= \text{DoubleConv}([\text{Up}(d_2),\ e_1],\ b)
    \label{eq:unet_dec}
\end{align}
where $[\cdot, \cdot]$ denotes channel-wise concatenation and $\text{Up}(\cdot)$ denotes a $2\times$ transposed convolution. A final $1 \times 1$ convolution maps $d_1$ to the single-channel predicted residual $\hat{r} = \text{Conv}_{1\times1}(d_1)$.

The skip connections are central to the U-Net's ability to model multi-scale residual structure: the bottleneck captures coarse, low-frequency error patterns introduced by the quantization bottleneck, while each decoder level, enriched by the corresponding encoder feature map, recovers progressively finer spatial detail. This is precisely what distinguishes our approach from the flat CNNs and tiny fully-connected models used in prior scientific residual predictors \cite{jia2025neurlz}.

\subsubsection{Training Objective}

The U-Net is trained as a separate, offline stage using pairs $(\tilde{x}, x)$ extracted from the RVQ output. The training target is the true pixel-space residual $r = x - \tilde{x}$, and we minimize the $\ell_1$ loss between the predicted and true residual:
\begin{equation}
    \mathcal{L}_{\text{U-Net}} = \frac{1}{n} \sum_{i=1}^{n} |\hat{r}_i - r_i|.
    \label{eq:unetloss}
\end{equation}
We use the $\ell_1$ norm rather $\ell_2$ following empirical evidence from the super-resolution literature \cite{lim2017enhanced} where it was shown that $\ell_1$ provides better convergence on structured data by treating more errors uniformly rather than over-penalizing rare large errors. The U-Net is trained with the Adam optimizer at a learning rate of $10^{-4}$, with a $70$/$30$ train/validation split.

\subsubsection{Refined Reconstruction}

After training, at inference time the U-Net takes $\tilde{x}$ as input and produces $\hat{r}$. The refined reconstruction is
\begin{equation}
    \tilde{x}_{\text{refined}} = \tilde{x} + \hat{r}.
    \label{eq:refined}
\end{equation}
This refined reconstruction is not guaranteed to meet the error bound $\tau$---the U-Net is a learned predictor that minimizes aggregate error rather than certifying any individual block. The role of Stage 3 is to provide this certificate.

\subsubsection{Bitrate Accounting}

Since the U-Net weights must be stored alongside the compressed bitstream to enable decompression, its parameter count contributes to the total bit cost. We did our experiments on two different U-Net sizes. With base channel count $b = 8$, our smaller 3-level U-Net has approximately $0.12$M parameters, contributing a fixed overhead of $\sim 0.46$~MB (stored at float32). While, our larger 3-level U-Net, with base channel count $b = 32$, has approximately $1.92$M parameters, contributing a fixed overhead of $\sim 7.35$~MB (stored at float32). This cost is amortized over the entire dataset being compressed and is negligible for large scientific volumes.

\subsection{Stage 3: Guaranteed Autoencoder (GAE)}

Even after U-Net correction, the refined reconstruction $\tilde{x}_{\text{refined}}$ may not satisfy the user-specified error bound $\tau$ at every spatial location. The GAE stage \cite{li2024attention} provides a deterministic, per-block guarantee by computing the remaining residual, projecting it onto a PCA basis, and retaining the minimum number of coefficients needed to meet the bound.

\subsubsection{Patch Vectorization}

The spatial frame $\tilde{x}_{\text{refined}}$ and the original $x$ are partitioned into non-overlapping $p \times p$ patches (we use $p = 8$, so each patch is a vector $v \in \mathbb{R}^{64}$). Let $\rho_i = x_i - \tilde{x}_{\text{refined},i}$ denote the residual vector for patch $i$. Patches whose residual norm already satisfies $\|\rho_i\|_2 \leq \tau\sqrt{p^2}$ are declared compliant and require no correction; only patches with $\|\rho_i\|_2 > \tau\sqrt{p^2}$ are processed by the GAE.

\subsubsection{PCA Basis Computation}

For the set of non-compliant patches, we fit a PCA basis $U \in \mathbb{R}^{M \times p^2}$ by computing the eigendecomposition of the empirical covariance matrix of the residual vectors. The projection coefficients for patch $i$ are
\begin{equation}
    c_i = U^\top \rho_i.
    \label{eq:pca_project}
\end{equation}
The coefficients are sorted in descending order of energy $c_{i,j}^2$, and we greedily select the minimum number of coefficients $M_i$ such that the remaining unexplained residual energy satisfies the bound
\begin{equation}
    \left\| \rho_i - U_{M_i} c_{i,M_i} \right\|_2 \leq \tau \sqrt{p^2}.
    \label{eq:pca_bound}
\end{equation}

\subsubsection{Quantization and Encoding}

Selected coefficients are quantized using a uniform bin width $b_q$. To preserve the error guarantee after quantization, the bin width is bounded by
\begin{equation}
    b_q \leq \frac{2\tau\sqrt{p^2}}{\sqrt{M_i}}
    \label{eq:binwidth}
\end{equation}
which follows from the fact that the worst-case $\ell_2$ error introduced by quantizing $M_i$ coefficients each by at most $b_q/2$ is $\frac{b_q}{2}\sqrt{M_i}$. The quantized coefficients, the PCA basis vectors, a binary mask indicating which coefficients were retained, and the mask lengths are then entropy-coded using GPU-accelerated lossless compression (nvcomp Zstandard) for efficient storage.

\subsubsection{Corrected Reconstruction}

The final GAE-corrected reconstruction for each patch is
\begin{equation}
    x^G_i = \tilde{x}_{\text{refined},i} + U_{M_i} c_{i,M_i}^q
    \label{eq:gae_recon}
\end{equation}
where $c_{i,M_i}^q$ denotes the quantized selected coefficients. By construction, $\text{NRMSE}(x_i, x^G_i) \leq \tau$ for every patch $i$, providing the per-block guarantee required by scientific applications.

\subsubsection{Role of the U-Net in Reducing GAE Cost}

The key insight motivating the placement of the U-Net before the GAE is that the GAE's compression efficiency depends directly on the magnitude of the residual it must correct. If $\tilde{x}_{\text{refined}}$ is already close to $x$ (i.e., the U-Net has absorbed much of the spatial error), then the residual $\rho_i = x_i - \tilde{x}_{\text{refined},i}$ is smaller in norm and lower in energy, allowing the GAE to meet the bound $\tau$ with fewer PCA coefficients $M_i$. Fewer coefficients directly translates to fewer bits in the compressed GAE stream, improving the overall compression ratio at the same $\tau$.

The complete pipeline proceeds as follows:

\begin{enumerate}
    \item \textbf{Tiling:} The input volume $(V, S, F, H, W)$ is tiled into spatial blocks of size $256 \times 256$ with reflect padding.
    \item \textbf{RVQ Compression (Stage 1):} Each tile is encoded by the VAE encoder, quantized with $N$ cascaded RVQ codebooks, and decoded to produce the pixel-space reconstruction $\tilde{x}$.
    \item \textbf{U-Net Residual Correction (Stage 2):} The U-Net takes $\tilde{x}$ as input and predicts the residual $\hat{r}$, producing the refined reconstruction $\tilde{x}_{\text{refined}} = \tilde{x} + \hat{r}$.
    \item \textbf{GAE Error Bounding (Stage 3):} The GAE computes the remaining residual $\rho = x - \tilde{x}_{\text{refined}}$, projects it onto a per-block PCA basis, and retains the minimum number of quantized coefficients needed to enforce $\text{NRMSE} \leq \tau$, producing the final certified output $x^G$.
    \item \textbf{Reassembly:} Tiles are reassembled by undoing the reflect padding, and normalization is inverted to restore the original data range.
\end{enumerate}

The total compressed bitstream consists of the RVQ model weights, the U-Net model weights, the PCA basis vectors, quantized coefficients, and binary selection masks from the GAE stage.

\section{Experiments and Results}
\label{sec:results}

\subsection{Datasets}

We evaluate our pipeline on three benchmark scientific simulation datasets that are standard in the learned scientific data compression literature \cite{li2025caesar}.

\textbf{E3SM} \cite{golaz2019doe} is a high-resolution Earth system model that simulates global atmospheric dynamics. We use hourly atmospheric data on a $0.25^\circ$ grid, yielding a dataset of shape $5 \times 6 \times 9360 \times 240 \times 240$ (variables $\times$ sections $\times$ frames $\times$ height $\times$ width).

\textbf{S3D} \cite{yoo2011direct} captures direct numerical simulations of compression ignition for a fuel-lean n-heptane/air mixture. The dataset covers 58 chemical species over 50 frames on a $640 \times 640$ spatial grid, forming tensors of shape $58 \times 1 \times 50 \times 640 \times 640$.

\textbf{JHTDB} \cite{li2008public} provides isotropic turbulence simulation data from the Johns Hopkins Turbulence Database. Our subset spans 64 spatial regions over 256 frames at $512 \times 512$ spatial resolution, with shape $3 \times 64 \times 256 \times 512 \times 512$.

These three datasets span diverse physical phenomena — climate, combustion, and fluid turbulence — and present very different spatial correlation structures, making them a rigorous testbed for our approach. All experiments are run on an NVIDIA B200 GPU on the HiPerGator supercomputer.

\subsection{Evaluation Metric}

We use Normalized Root Mean Square Error (NRMSE) as our primary reconstruction quality metric, consistent with prior work \cite{li2025caesar, li2024attention}:
\begin{equation}
    \text{NRMSE}(\Omega, \hat{\Omega}) = \frac{\sqrt{\frac{1}{N_d}\|\Omega - \hat{\Omega}\|_2^2}}{x_{\max} - x_{\min}}
    \label{eq:nrmse_results}
\end{equation}
where $N_d$ is the total number of data points and $x_{\max} - x_{\min}$ is the value range of the original data. We report final NRMSE after the complete pipeline (RVQ + U-Net + GAE) as a function of Compression Ratio (CR), where $\text{CR} = \frac{32 \cdot N_d}{B_{\text{total}}}$ with $B_{\text{total}}$ the total number of bits in the compressed bitstream.

\subsection{Implementation Details}

\subsubsection{Stage 1: RVQ-VAE Training}

The RVQ-VAE encoder-decoder uses a ResNet-based 2D architecture with dimension multipliers $[1, 2, 3, 4]$ (encoding) and $[4, 3, 2, 1]$ (decoding), with base model dimension $d = 16$ and single-channel input and output. The RVQ quantizer uses $N = 4$ cascaded codebooks, each of size $K = 64$. The EMA decay for codebook updates is $\gamma = 0.99$ and the commitment loss weight is $0.25$. The model is trained with the Adam optimizer at a learning rate of $5 \times 10^{-4}$, with a MultiStep learning rate schedule that halves the rate at $20\%$, $40\%$, $60\%$, and $80\%$ of total training, for a total of $700{,}000$ iterations with batch size $64$. Training uses MSE loss augmented by the commitment loss from the RVQ stages. Each dataset is trained independently. The model size is $1.290$~MB (stored at float32).

\subsubsection{Stage 2: U-Net Training}

The U-Net is trained as an independent post-processing stage, taking the saved RVQ reconstructions as input. The architecture uses 3 encoder levels with base channel counts $b \in \{8, 32\}$, yielding two model variants that we denote \textbf{U-Net (small)} ($b = 8$, $\approx 0.12$M parameters, $0.462$~MB) and \textbf{U-Net (large)} ($b = 32$, $\approx 1.92$M parameters, $7.349$~MB after additional overhead). The training uses the AdamW optimizer with learning rate $5 \times 10^{-3}$ for \textbf{U-Net (small)} and $5 \times 10^{-4}$ for \textbf{U-Net (large)}; and weight decay $10^{-4}$, with a Cosine Annealing schedule over up to $10{,}000$ epochs and batch size $512$. The loss function is $\ell_1$ on the predicted residual. The data is split 70/15/15 into train/validation/test sets, and the best checkpoint is selected by minimum validation $\ell_1$ loss. Spatial tiles of $256 \times 256$ are processed frame-by-frame as independent samples.

All three stages are trained and evaluated in a dataset-specific compression setting. For each dataset (E3SM, S3D, JHTDB), the RVQ-VAE is trained independently on frames drawn from that dataset. The U-Net is subsequently trained on the RVQ reconstruction residuals from the same dataset, using a 70/15/15 train/validation/test split applied at the frame level. The reported NRMSE vs.\ CR curves reflect performance on the overall dataset once our U-Net model is ready and during inference. The GAE is applied at inference time and requires no training; it adapts per-tile at compression time. This setting evaluates dataset-specific compression performance, i.e., how well the pipeline compresses data from a fixed known domain, which is the standard evaluation protocol in the scientific data compression literature \cite{li2025caesar, jia2025neurlz}.

\subsubsection{Stage 3: GAE}

The GAE uses $8 \times 8$ non-overlapping patches (patch vectors of dimension $p^2 = 64$). PCA is fit on the full set of residual vectors per spatial tile. Coefficients are quantized with the bin width upper bound from \autoref{eq:binwidth}, and the final compressed stream is entropy-coded with Zstandard via GPU-accelerated nvcomp. We sweep NRMSE target bounds $\tau \in \{0.0001, 0.0002, 0.0005, 0.001, 0.002, 0.004\}$ (and additional values per dataset as visible in the results plots) to obtain the full NRMSE vs.\ CR trade-off curve.

\subsection{Baselines}

We compare against two baselines. The primary baseline is \textbf{RVQ only} (Stage 1 alone, followed directly by GAE with no U-Net correction), which isolates the contribution of the U-Net stage. We also include results for \textbf{U-Net Residual (L1)}, which denotes a setting where the U-Net is used as a residual predictor before the GAE is applied to the refined RVQ output with the U-Net correction; as presented in \autoref{fig:s3d_losses} for the ablation discussion.

\subsection{Main Results: NRMSE vs. Compression Ratio}

\begin{figure}[!htbp]
    \centering
    \includegraphics[width=0.7\linewidth]{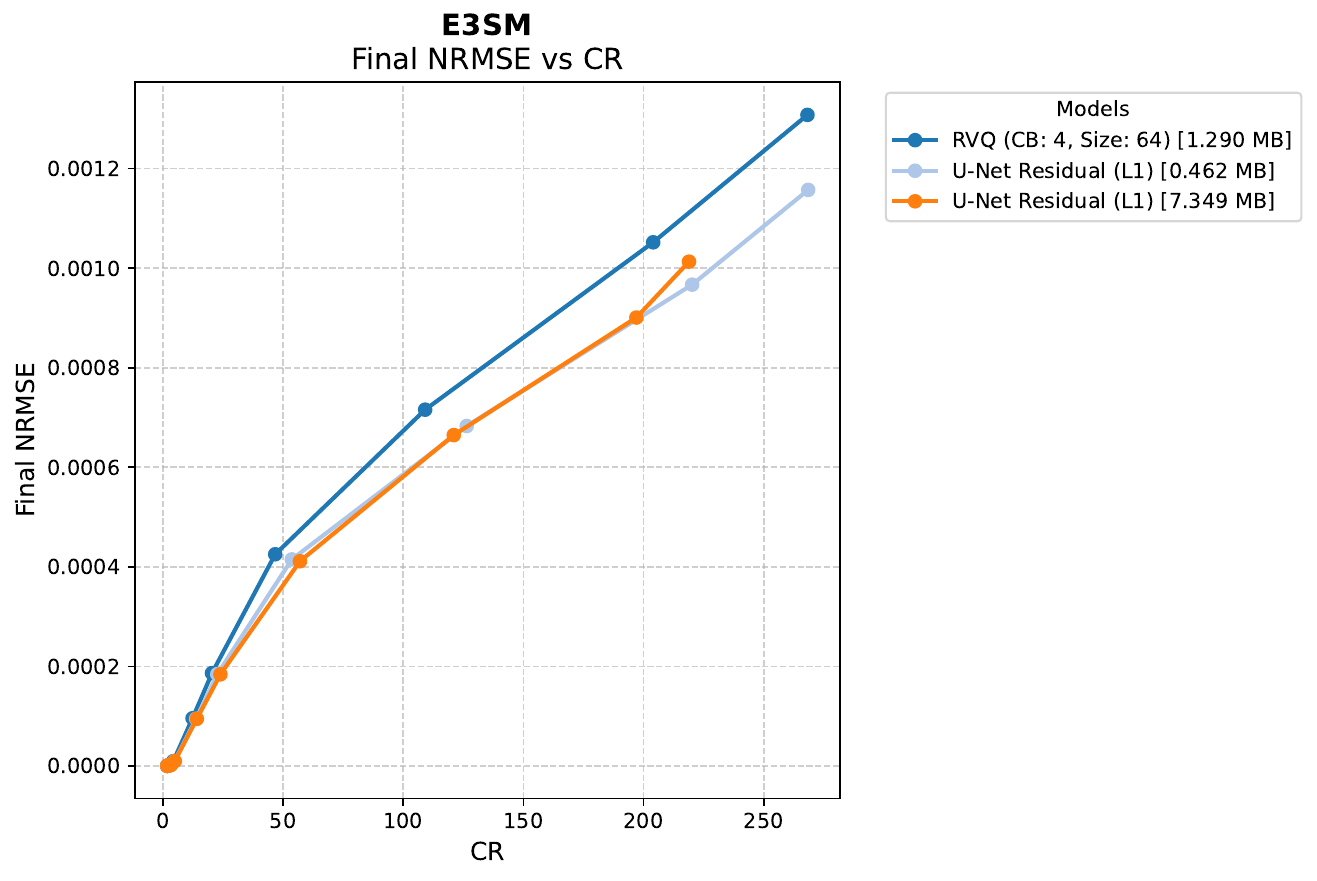}
    \caption{Final NRMSE vs.\ Compression Ratio on the E3SM dataset. RVQ (CB: 4, Size: 64) is the Stage 1 baseline. U-Net Residual (L1) with 0.462~MB and 7.349~MB model sizes correspond to the small ($b=8$) and large ($b=32$) U-Net variants respectively. The U-Net correction consistently shifts the curve downward, achieving lower NRMSE at the same compression ratio across the full operating range.}
    \label{fig:e3sm_results}
\end{figure}

\begin{figure}[!htbp]
    \centering
    \includegraphics[width=0.7\linewidth]{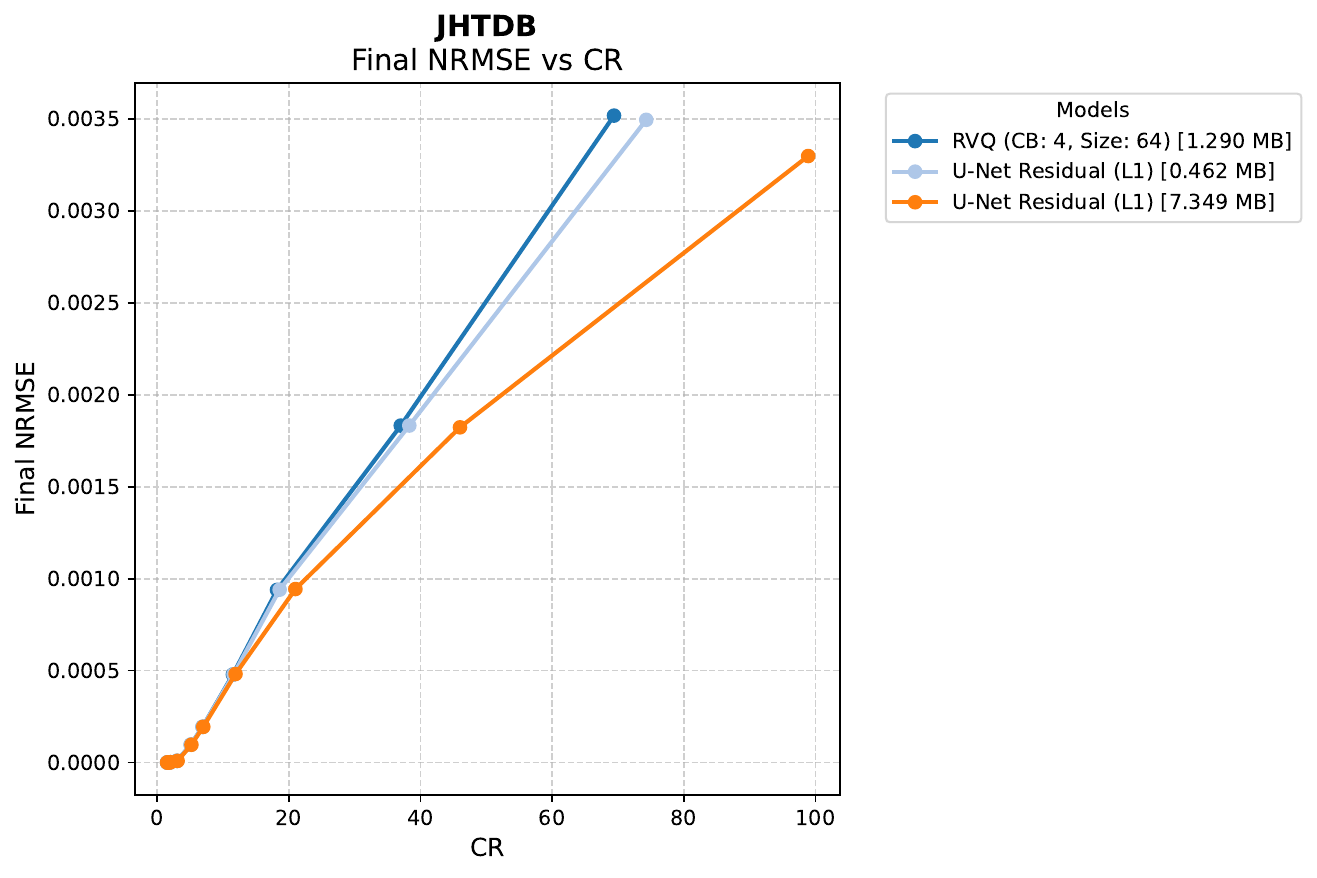}
    \caption{Final NRMSE vs.\ Compression Ratio on the JHTDB dataset. The turbulence dataset presents the most structured and high-frequency residuals among our three benchmarks. Both U-Net variants improve over RVQ alone, with the larger model providing a more consistent gain especially at high compression ratios.}
    \label{fig:jhtdb_results}
\end{figure}

\begin{figure}[!htbp]
    \centering
    \includegraphics[width=0.7\linewidth]{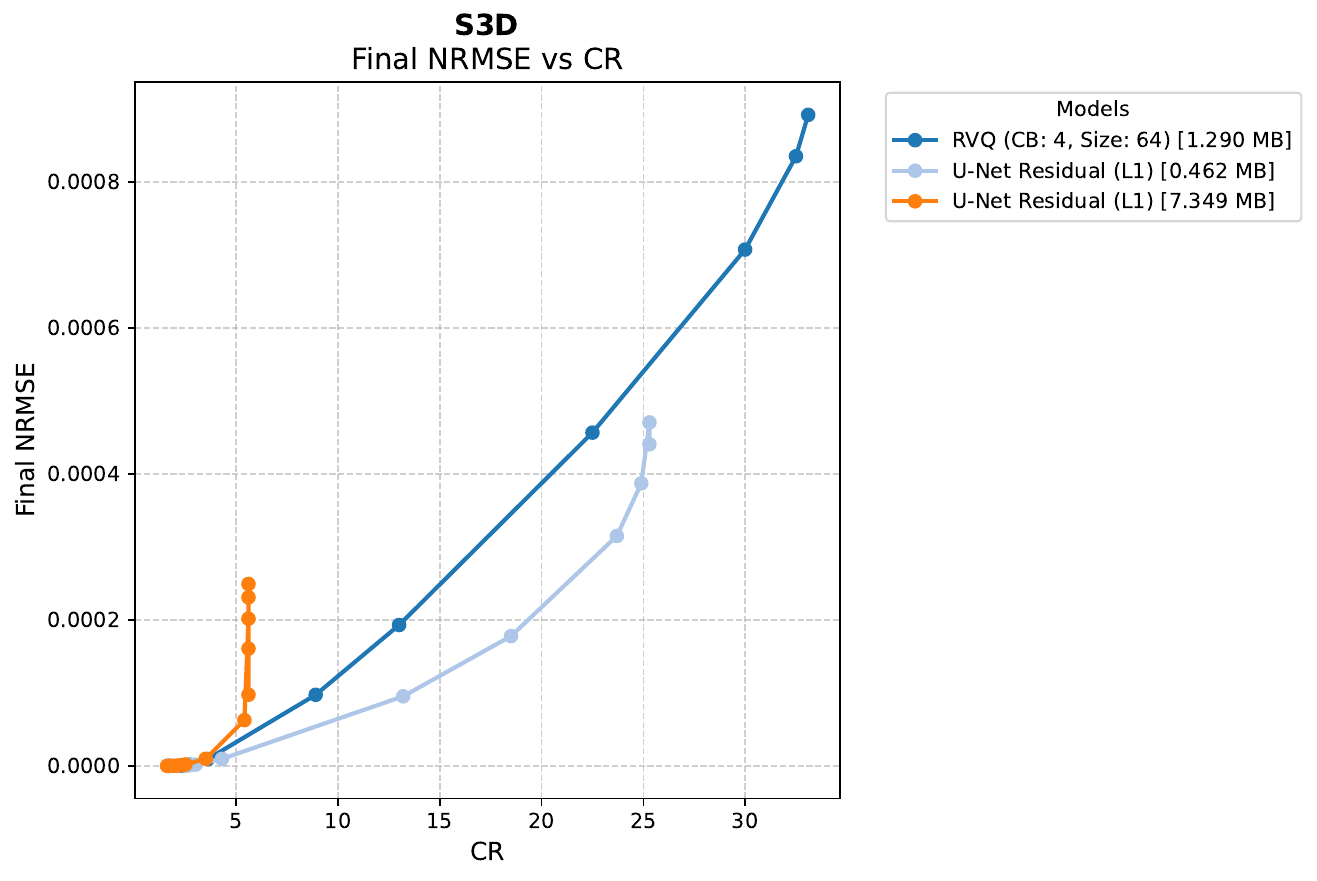}
    \caption{Final NRMSE vs.\ Compression Ratio on the S3D dataset. The smaller U-Net variant improves over the RVQ baseline with the larger model achieving a more substantial gain in NRMSE. But we take a huge hit in the compression ratios. This highlights a case where the U-Net model size matters since we are working with a comparatively smaller dataset.}
    \label{fig:s3d_results}
\end{figure}

\autoref{fig:e3sm_results}, \ref{fig:s3d_results}, and \ref{fig:jhtdb_results} show the final NRMSE vs.\ CR curves for E3SM, S3D, and JHTDB respectively, comparing the RVQ-only baseline against both U-Net variants with $\ell_1$ loss. Here final NRMSE is not same as the Target NRMSE. Final NRMSE is the NRMSE got after setting the target bound. It is less than or max equal to Target NRMSE.

Across all three datasets, the U-Net correction stage consistently and substantially improves upon the RVQ-only baseline at every compression ratio. On E3SM, which has relatively smooth atmospheric fields, both U-Net variants provide consistent improvement across the full CR range, with the larger model slightly outperforming the smaller one. The gain is especially pronounced at mid-to-high CRs (above $\approx 100$), where RVQ quantization error is large and the U-Net has more structured residual signal to recover. On JHTDB, the turbulence dataset with the most fine-scale spatial structure, the U-Net provides substantial improvements throughout the CR range. On S3D, as shown in \autoref{fig:s3d_results}, which contains sharp chemical species gradients, the improvement is particularly pronounced at high compression ratios for the smaller U-Net. But it is also an important case which shows how U-Net weights matter in terms of compression ratio. As this dataset is very small the U-Net weights have a significant impact in the bitstream amount as a result the larger U-Net which performed better in the other two datasets, also gives better NRMSE here but at the cost of compression ratio as its size is increasing the overall bits significantly.

These results directly support the central claim of our pipeline: RVQ residual modeling in latent space leaves a pixel-space reconstruction error that has spatial structure which a dedicated post-processing network can exploit to improve compression efficiency. The improvement is not merely a bitrate shift; because the U-Net correction reduces the magnitude of the residual seen by the GAE, the GAE requires fewer PCA coefficients to enforce the same NRMSE bound $\tau$, resulting in a genuine compression ratio gain at the same reconstruction quality.

\subsection{Effect of U-Net Model Size}

A notable observation across all three datasets is that the two U-Net variants — small ($0.462$~MB) and large ($7.349$~MB) — both improve over the RVQ baseline, though the larger model provides a more consistent and often larger gain. This is an important practical finding: even a very lightweight U-Net ($b = 8$, roughly $0.12$M parameters) provides meaningful pixel-space residual correction. Because the U-Net weights are stored once as a fixed overhead in the compressed bitstream and amortized over the entire dataset, the cost of the larger model is relatively modest for large scientific volumes. We do not fix the U-Net size as a design constraint — it is a hyperparameter that can be tuned based on the available storage budget and the size of the dataset being compressed.

\subsection{Ablation: Effect of U-Net Loss Function on S3D}

\begin{figure}[!htbp]
    \centering
    \includegraphics[width=0.7\linewidth]{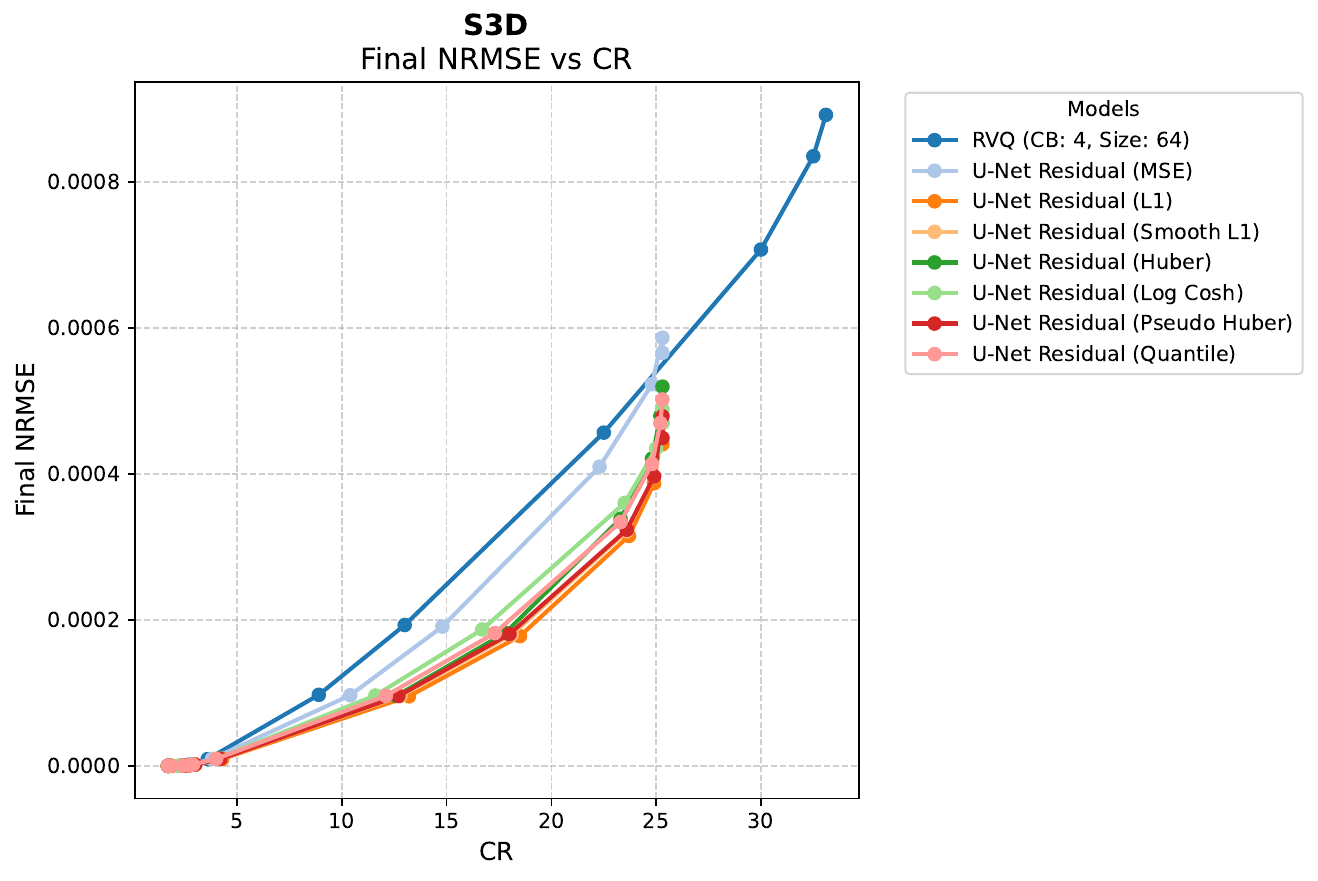}
    \caption{Ablation study of U-Net loss functions on the S3D dataset. All loss variants substantially outperform the RVQ-only baseline. Robust losses like $\ell_1$, Smooth L1, Huber, Log-Cosh, and Pseudo-Huber, cluster tightly together and outperform MSE, suggesting that residual-robust loss functions are more appropriate for the heavy-tailed residual distribution typical of learned scientific data compressors.}
    \label{fig:s3d_losses}
\end{figure}

\autoref{fig:s3d_losses} presents an ablation study on the S3D dataset comparing seven loss functions for the smaller U-Net residual predictor: MSE, $\ell_1$, Smooth L1, Huber, Log-Cosh, Pseudo-Huber, and Quantile loss.

Several conclusions emerge from this ablation. First, all U-Net variants substantially improve over the RVQ baseline and confirming that the U-Net stage provides a consistent benefit independent of the choice of loss. Second, MSE performs slightly worse than the robust losses ($\ell_1$, Smooth L1, Huber, Log-Cosh, Pseudo-Huber), which cluster tightly together. This is consistent with the observation in the super-resolution literature \cite{lim2017enhanced} that $\ell_1$ tends to be more stable than MSE on structured reconstruction tasks: the compression residual of a scientific VAE has a heavy-tailed distribution (a few pixels with large error, many pixels near zero), and $\ell_1$ weights all residual magnitudes uniformly rather than over-penalizing the rare large errors that MSE amplifies. Based on these results, we adopt $\ell_1$ as the default loss for all other experiments.

\subsection{Qualitative Results: Per-Tile Visualization}

\begin{figure}[!htbp]
    \centering
    \includegraphics[width=\linewidth]{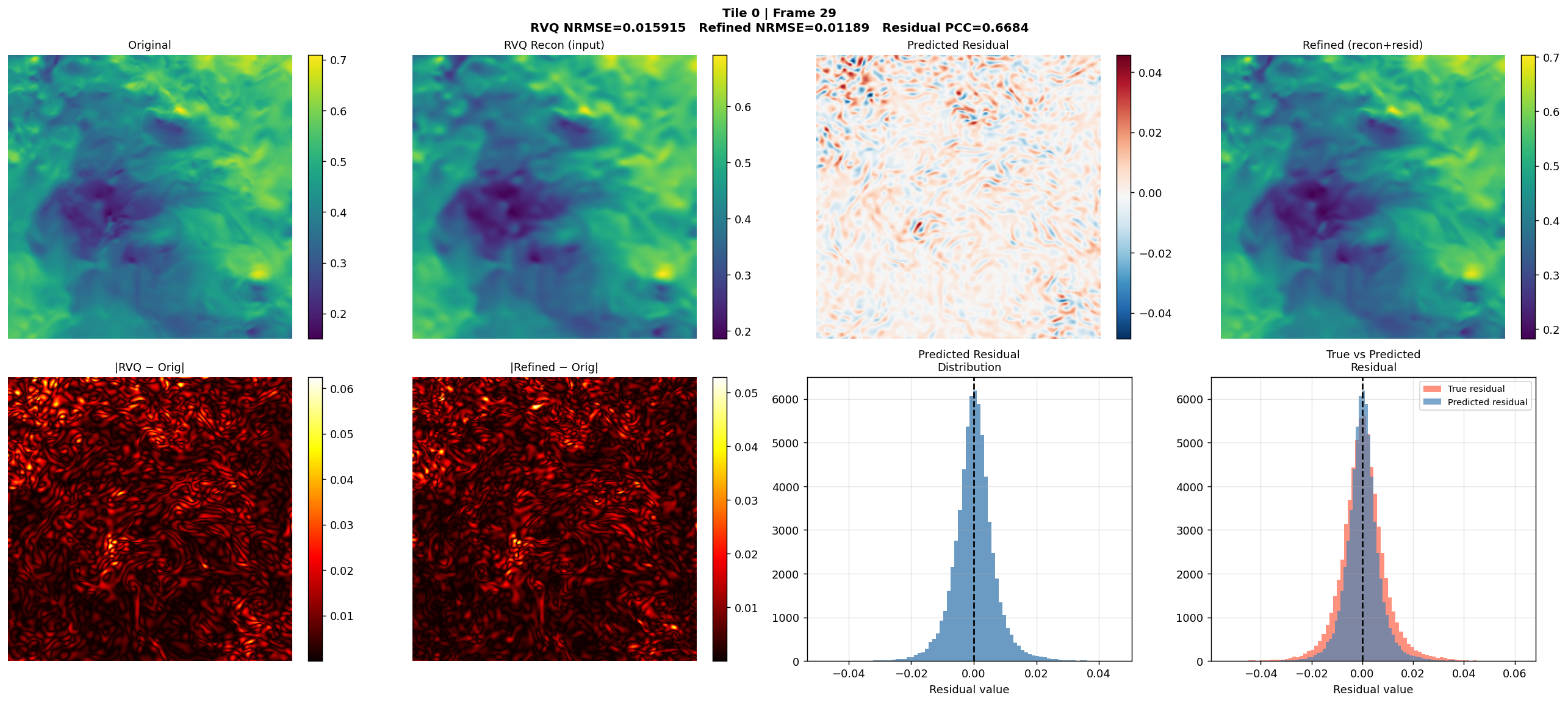}
    \caption{Qualitative visualization for a sample representative JHTDB tile and frame. Top row (left to right): original field, RVQ reconstruction, predicted residual $\hat{r}$, refined reconstruction $\tilde{x}_{\text{refined}} = \tilde{x} + \hat{r}$. Bottom row: absolute error map $|$RVQ $-$ Original$|$, absolute error map $|$Refined $-$ Original$|$, predicted residual histogram, and true vs.\ predicted residual overlay. The U-Net substantially reduces the absolute error (visible as reduced intensity in the error map), and the predicted residual distribution closely matches the true residual distribution, confirming that the U-Net has learned a useful approximation to the pixel-space reconstruction error.}
    \label{fig:per_tile}
\end{figure}

\autoref{fig:per_tile} shows a sample representative per-tile qualitative comparison on the JHTDB dataset, displaying the original field, the RVQ reconstruction, the predicted residual, and the final refined reconstruction, alongside per-pixel absolute error maps and residual histograms.

The qualitative results make the effect of our pipeline tangible. The RVQ reconstruction retains the broad spatial structure of the original field but introduces fine-scale errors particularly along flow boundaries and high-gradient regions. The predicted residual captures precisely these spatially structured error patterns, concentrated along the same high-frequency structures, demonstrating that the U-Net has learned to model the spatial topology of the quantization error, not just its magnitude. The refined reconstruction visibly reduces these artifacts. The residual histograms confirm that the predicted residual distribution shows meaningful alignment with the true residual distribution, with a Pearson Correlation Coefficient (PCC) of $0.6684$ for this tile and frame, indicating that the U-Net captures a substantial portion of the spatial residual structure while some prediction error remains.

\subsection{Bitstream Decomposition}

A key motivation for our pipeline is that the U-Net correction reduces the bitrate required by the GAE correction stream. \autoref{tab:bitstream} provides a breakdown of the total compressed bitstream for a representative compression point on each dataset, illustrating how the bits are distributed across the four components of our pipeline.

\begin{table}[!htbp]
\caption{Approximate bitstream decomposition for a representative compression point on each dataset. RVQ indices \& codebook: bits used by the Huffman entropy-coded quantized codebook assignments and the stored codebook vectors across all $N=4$ RVQ stages. RVQ model: one-time cost of storing the encoder-decoder weights. U-Net model: one-time cost of storing the residual predictor weights. GAE bits: bits used by the GAE to store coefficients, etc. for both the U-Net variants and RVQ-only (For Target NRMSE: 0.002). All values are approximate and vary with the NRMSE target.}
\label{tab:bitstream}
\centering
\begin{adjustbox}{width=\textwidth}
\begin{tabular}{l|ccc|c}
\hline
\textbf{Component} & \textbf{E3SM} & \textbf{S3D} & \textbf{JHTDB} & \textbf{Notes} \\
\hline
Dataset Size & 7.3 GB & 50 MB & 6 GB & Sizes of the dataset chunk we used\\
RVQ indices \& codebooks & 25.58 MB & 0.215 MB & 18.86 MB & Huffman entropy-coded indices \\
RVQ model weights & 1.290 MB & 1.290 MB & 1.290 MB & Fixed overhead \\
U-Net model weights & 0.462 / 7.349 MB & 0.462 / 7.349 MB & 0.462 / 7.349 MB & Small / Large \\
GAE PCA bits (U-Net small) & 7.047 MB & 0.023 MB & 140.7 MB & Reduced by U-Net small \\
GAE PCA bits (U-Net large) & 4.226 MB & 0.018 MB & 107.0 MB & Reduced by U-Net large \\
GAE PCA bits (RVQ only) & 10.24 MB & 0.042 MB & 146.7 MB & Bits required after RVQ-only \\
\hline
\end{tabular}
\end{adjustbox}
\end{table}

We draw your attention to the GAE PCA coefficients row: because the U-Net pre-corrects the pixel-space residual, the residual $\rho = x - \tilde{x}_{\text{refined}}$ that the GAE must project is smaller in both magnitude and spatial complexity than $\rho = x - \tilde{x}$ without the U-Net. This directly reduces the number of PCA bits $M_i$ needed per patch to satisfy the same bound $\tau$, which is the mechanism by which the U-Net improves the NRMSE vs.\ CR curve. The RVQ model and U-Net model costs are one-time fixed overheads that are amortized over the entire dataset and become negligible for the large volumes typical of scientific simulations.

\subsection{Discussion}

The results consistently demonstrate that the U-Net pixel-space residual correction stage provides a reliable, dataset-agnostic improvement over RVQ-only compression followed by GAE. The improvement is present across all three physically diverse datasets, is robust to the choice of loss function (with $\ell_1$ and related robust losses performing best), and scales with model capacity in a predictable way. The qualitative visualizations confirm that the U-Net has learned a meaningful spatial model of the RVQ reconstruction error rather than simply adding noise or a constant offset. Together, these results validate the central hypothesis of our work: RVQ's latent-space residual correction does not address the pixel-space reconstruction error, and a dedicated pixel-space neural post-processor can exploit the spatial structure of this error to improve compression efficiency under a guaranteed error bound.

\subsection{Generalizability: U-Net as a Post-Processor for Traditional Compressors}

To evaluate whether the pixel-space residual correction benefit of our U-Net stage generalizes beyond the RVQ-based pipeline, we conduct an additional experiment on the S3D dataset using SZ3 \cite{liang2022sz3} as the base compressor in place of the RVQ. SZ3 is one of the most widely adopted error-bounded lossy compressors for scientific data, and serves here as a strong traditional baseline.

We compress the S3D dataset with SZ3 at a target NRMSE of $1 \times 10^{-3}$, obtaining a compression ratio of $29.52$ at a final NRMSE of $9.96 \times 10^{-4}$. We then attach our pre-trained U-Net (small variant, $b=8$) as a post-processing stage on top of SZ3's reconstruction, exactly as we do in the primary pipeline. The results are summarized in \autoref{tab:sz_unet}.

\begin{table}[!htbp]
\caption{Comparison of SZ3 alone vs.\ SZ3 + U-Net (small) on the S3D dataset at target NRMSE $= 1 \times 10^{-3}$. The U-Net model overhead is included in the total bitstream for the SZ3 + U-Net case.}
\label{tab:sz_unet}
\centering
    \begin{tabular}{lccc}
    \hline
    \textbf{Method} & \textbf{NRMSE} & \textbf{CR} \\
    \hline
    SZ3 only          & $9.956 \times 10^{-4}$ & 29.52 \\
    SZ3 + U-Net (small) & $6.163 \times 10^{-4}$ & 12.99 \\
    \hline
    \end{tabular}
\end{table}

The U-Net correction reduces the final NRMSE by approximately $38\%$ relative to SZ3 alone ($9.96 \times 10^{-4} \rightarrow 6.16 \times 10^{-4}$), confirming that the pixel-space residual structure left behind by a traditional predictor-quantization compressor is also amenable to neural correction, not just that of a learned VAE. However, the compression ratio drops from $29.52$ to $12.99$, because the U-Net model weights ($0.462$~MB, small variant) constitute a fixed overhead that is non-negligible relative to SZ3's already compact bitstream on this small experimental subset of S3D.

This trade-off is inherently dataset-size dependent. The U-Net model cost is a one-time fixed overhead that is amortized over the entire volume being compressed. For large-scale scientific datasets such as the full E3SM ($61$~GB) or JHTDB ($45$~GB), the $0.462$~MB model overhead becomes negligible — less than $0.001\%$ of the raw data size — and the NRMSE improvement would translate directly into a compression ratio gain rather than a loss. The S3D subset used here is small enough that the model overhead dominates, but this is a property of the dataset size, not of the approach itself.

These results suggest that our U-Net residual correction stage is broadly applicable as a modular post-processor for scientific data compressors in general, not just for RVQ-based learned pipelines, opening up an avenue for improving traditional compressors such as SZ3 and ZFP \cite{lindstrom2014fixed} with minimal architectural modification.

\section{Conclusion}\label{sec:conclusion}

In this work, we presented a three-stage post-processing pipeline for improving scientific data compression, built on top of an RVQ-based VAE compressor. The central observation motivating our approach is that while RVQ performs residual correction in the latent space of the compressor, the pixel-space reconstruction error that remains after decoding is spatially structured and cannot be addressed by latent-space quantization alone. We showed, through correlation analysis on the JHTDB dataset, that this residual is not driven by simple local statistics such as gradient magnitude or local variance, motivating the use of a deep spatial model for its prediction.

Our pipeline addresses this gap by inserting a dedicated pixel-space U-Net residual predictor between the RVQ decoder and the Guaranteed Autoencoder (GAE) error-bound enforcer. The U-Net takes the RVQ reconstruction as input, predicts the pixel-space residual, and produces a refined reconstruction whose remaining error is both smaller in magnitude and simpler in spatial structure than the original RVQ residual. This directly reduces the number of PCA coefficients the GAE must store to enforce a given NRMSE bound $\tau$, improving the overall compression ratio at the same reconstruction quality. The GAE then provides a hard per-block guarantee that the final output meets $\tau$ regardless of how well the upstream stages perform.

We evaluated our pipeline on three diverse scientific simulation datasets, namely E3SM (climate), S3D (combustion), and JHTDB (fluid turbulence); and demonstrated consistent improvements in NRMSE vs.\ compression ratio across all three. The improvement holds across a range of U-Net loss functions, with robust losses ($\ell_1$, Smooth L1, Huber) outperforming MSE, and scales predictably with U-Net model capacity. Even a lightweight U-Net with $\approx 0.12$M parameters provides meaningful gains, making the approach practical for large scientific volumes where the fixed model overhead is amortized over the dataset.

\subsection{Future Work}

Several directions remain open for investigation and represent natural extensions of the current pipeline.

\textbf{Topological supervision via persistence landscapes.} The current U-Net is trained with pixel-wise $\ell_1$ loss, which penalizes pointwise reconstruction error but has no mechanism to preserve the topological structure of the scalar field i.e., features such as connected components, loops, and voids that are critical for downstream scientific analysis such as vortex identification and combustion front tracking. A natural extension is to augment the U-Net training objective with a topological loss term based on persistence landscapes \cite{bubenik2015statistical}. Persistence landscapes are stable, differentiable representations of the topological features of a scalar field, computed via persistent homology. Preliminary work in our group has shown that RVQ preserves the first persistence landscape $\lambda_1$ reasonably well but introduces hundreds of spurious low-persistence pairs visible in higher landscape layers $\lambda_2, \lambda_3, \ldots$, suggesting that a topological supervision signal targeting these higher layers could guide the U-Net to suppress spurious features while recovering genuine topological structure lost during quantization. This would make the pipeline sensitive not just to pointwise fidelity but to the preservation of physically meaningful spatial topology.

\textbf{Guaranteed Topological Error (GTE) module.} Analogous to the GAE's hard per-block NRMSE guarantee, a Guaranteed Topological Error module could enforce a user-specified bound on the topological distance between the original and reconstructed scalar fields, measured via the persistence landscape norm $\|\lambda_k(x) - \lambda_k(\hat{x})\|_2$. This would extend the pipeline's error-bounding capability from purely quantitative (NRMSE) to structurally meaningful (topology), addressing a key limitation of all current learned scientific compressors.

\textbf{Extension to three-dimensional volumetric fields.} The current pipeline processes scientific data frame-by-frame as independent 2D spatial tiles. A natural generalization is to replace the 2D U-Net with a 3D encoder-decoder that operates directly on volumetric blocks of shape $(h_b \times w_b \times d_b)$, capturing correlations across the depth or section dimension in addition to the spatial dimensions. This is particularly relevant for datasets such as JHTDB, where the velocity field is fully three-dimensional and frame-to-frame correlations carry significant physical information.

\textbf{Adaptive U-Net capacity and online adaptation.} The current pipeline trains a single U-Net per dataset offline. A promising direction is to explore online or meta-learning approaches that adapt the U-Net weights at compression time to the specific statistical properties of each data block, similar in spirit to NeurLZ \cite{jia2025neurlz} but with a full encoder-decoder architecture rather than a tiny fully-connected network. This could allow the model to specialize to locally heterogeneous residual distributions.

\textbf{Joint end-to-end training.} In the current pipeline, the RVQ-VAE, U-Net, and GAE are trained as independent stages. While this modular design has practical advantages, each stage can be developed, debugged, and replaced independently; it leaves potential gains on the table, since the RVQ decoder is never aware that a U-Net will correct its output. An end-to-end training strategy that jointly optimizes the RVQ and U-Net under the GAE's error-bound constraint could allow the RVQ to learn to concentrate its reconstruction error in regions where the U-Net is most effective, potentially improving both compression ratio and reconstruction quality beyond what the staged approach achieves.

\textbf{Broader scientific domains.} Our current evaluation covers climate, combustion, and fluid turbulence. Extending the pipeline to other scientific domains would test the generality of the residual-correction approach and may reveal domain-specific residual structures that call for architectural variations in the U-Net stage.

\section*{Acknowledgments}
The work of S.M, S.R and A.R was supported in part by the U.S. Department of Energy, Office of Science, Office of Advanced Scientific Computing Research's Computer Science Competitive Portfolios program under Contract No. DE-AC05-00OR22725.

\bibliographystyle{IEEEtran}
\bibliography{references}

\end{document}